# Recovering Individual-Level Activity Sequences from Location-Based Service Data Using a Novel Transformer-Based Model

Weiyu Luo
Department of Civil & Environmental Engineering
Villanova University, Villanova, PA 19085
Email: wluo01@villanova.edu

Chenfeng Xiong*
Department of Civil & Environmental Engineering
Villanova University, Villanova, PA 19085
Email: chenfeng.xiong@villanova.edu

Weiyu Luo, Chenfeng Xiong

**ABSTRACT**
Location-Based Service (LBS) data provides critical insights into human mobility, yet its sparsity often yields incomplete trip and activity sequences, making accurate inferences about trips and activities difficult. We raised a research problem: *Can we use activity sequences derived from high-quality LBS data to recover incomplete activity sequences at individual level?* This study proposes a new solution, the Variable Selection Network-fused Insertion Transformer (VSNIT), integrating the Insertion Transformer's flexible sequence construction with the Variable Selection Network's dynamic covariate handling capability, to recover missing segments in incomplete activity sequences while preserving existing data. The findings show that VSNIT inserts more diverse, realistic activity patterns, more closely matching real-world variability, and restores disrupted activity transitions more effectively aligning with the target. It also performs significantly better than the baseline model across all metrics. These results highlight VSNIT's superior accuracy and diversity in activity sequence recovery tasks, demonstrating its potential to enhance LBS data utility for mobility analysis. This approach offers a promising framework for future location-based research and applications.

**Keywords:** Sequence-To-Sequence Modeling, Location-Based-Service Data, Data Sparsity, Insertion Transformer, Activity-Based Modeling, Human Mobility

Weiyu Luo, Chenfeng Xiong

# INTRODUCTION
## Activity-based model
Activity-based modeling (ABM) emerged in response to the limitations of traditional trip-based models, providing a more behaviorally appropriate framework for understanding travel demand (*1–3*). Its conceptual roots can be traced back to the 1970s and 1980s, when researchers began to emphasize the relationship between travel and activity participation. This paradigm shift recognized that travel was not a final purpose in itself, but a derived need driven by the need to engage in various activities. In the ensuing decades, advances in computing power, data collection, and behavioral research have facilitated the development and application of ABM, resulting in a richer presentation of the decision-making processes of individuals and families. By focusing on activity sequences and their interdependencies, ABM has become a reliable alternative for modeling more complex travel behavior, especially in modern multimodal transportation systems.

The activity-based model shows several advantages over its trip-based predecessor. By emphasizing activity patterns rather than isolated trips, ABMs capture the interrelationships between activities, travel, and time constraints, providing a more comprehensive understanding of travel behavior. They account for phenomena such as trip chaining, activity prioritization, and the impacts of time-of-day decisions. Furthermore, ABMs incorporate individual and household-level heterogeneity, allowing for a more nuanced analysis of diverse travel behaviors and policy impacts.

Arentze and Timmermans (2004) proposed a learning-based transportation oriented simulation system (ALBATROSS) (*4*). This model integrates bounded rationality theory, assuming individuals use heuristics decision-making rather than optimizing all choices under utility maximization theory. It operates on decision trees to capture how individuals incrementally make scheduling decisions based on past experiences.

Kitamura and Fujii (1998) proposed a computational process model—Prism-Constrained Activity-Travel Simulator (PCATS) (*5*). PCATS incorporates Hägerstrand's time-space prism theory (*6*, *7*), modeling activity engagement and travel choices while considering temporal and spatial constraints. It decomposes activity-travel pattern into conditional probabilities and sequentially simulates activity type, location, travel mode and duration. The conditional probabilities are determined by utility maximization. PCATS has been subsequently applied to the FAMOS project in Florida in 2004 (*8*).

Pendyala et al. (1997) proposed a microsimulation model—Activity-Mobility Simulator (AMOS) (*9*) which is an adaptive behavioral model designed to simulate individuals' responses to changes in travel environments, such as congestion pricing or transportation demand management (TDM) policies. The model first receives an individual's baseline activity-travel patterns, which is then fed into a neural network to generate a response. Subsequently, a pattern modifier then intervenes to adjust the baseline pattern. The utility of the modified pattern is then evaluated and a satisficing rule is applied to decide whether to accept the modified pattern or not. AMOS has been subsequently embedded into a dynamic forecasting system called SAMS (*10*) for predicting transportation, land use and air quality.

Miller and Roorda (2003) introduced the Toronto Area Scheduling Model for Household Agents (TASHA) (*11*), a rule-based microsimulation model designed to generate daily activity-travel pattern for household. TASHA adopts the concept of "project" to contain and organize activity and travel episodes within household and person schedules. Activity and travel episodes are then inserted in the project dynamically. The attributes of the episodes are generated from empirical distributions.

Recker, McNally, and Root (1986) present an operational model of complex travel behavior, STARCHILD (*12*) which builds upon a theoretical framework developed in a preceding study. The model system integrates multiple methodological approaches, including simulation, pattern recognition, multi-objective optimization, and disaggregate choice models, to analyze the formation of household travel and activity patterns. A key assumption underlying the model is that travel decisions emerge from household-level constraints and interactions, rather than being strictly individualistic. The model incorporates five modules: (1) household activity program generation, (2) a combinatorial scheduling algorithm to produce feasible activity sequences, (3) pattern classification to reduce the choice set, (4) a multi-objective decision-making model to refine alternatives, and (5) a multinomial logit model for final activity pattern choice. The highlights of the model include its ability to account for intra-household interactions, its dynamic consideration of travel mode and time constraints, and its emphasis on individual decision-making within a constrained environment. Empirical validation demonstrates that STARCHILD successfully replicates observed travel behaviors, offering a robust tool for transportation policy analysis and planning.

Despite many strengths of ABMs, they are not without limitations. The implementation requires extensive data on activity patterns, individual constraints, and preferences. These data are largely collected through extensive surveys, which are challenging and costly to collect and process. Additionally, the computational complexity of ABMs necessitates significant resources for model development, calibration, and validation. These challenges, while substantial, have spurred ongoing advancements in data collection techniques, such as GPS and smartphone tracking, as well as the integration of machine learning algorithms to enhance the accuracy and scalability of activity-based approaches.

**Sequence-to-sequence model**
Many real-world problems can be abstracted as sequence-to-sequence (seq2seq) problems (*13*), e.g., machine translation, speech recognition, time-series forecasting, etc. Sequence modeling technique has been widely applied to solve these problems. In the past decade, sequence modeling has undergone significant evolution, particularly with deep learning advancements. Early sequence modeling relied on Recurrent Neural Networks (RNNs) (*14*), notably Long Short-Term Memory (LSTM) networks (*15*), which introduced gating mechanisms to address exploding or vanishing gradients in long-range dependencies. Further improvements were made with Gated Recurrent Units (GRUs) (*16*, *17*), which simplified LSTM's architecture by merging the forget and input gates. These models excelled in tasks like speech recognition and machine translation but still suffered from limited parallelization and difficulties capturing very long-term dependencies.

A major breakthrough came with the introduction of the Transformer model (*18*), which replaced recurrence with a self-attention mechanism. This design allowed for full parallelization during training and captured global dependencies effectively. Key to this success was the scaled dot-product attention mechanism, which computed attention scores across all tokens simultaneously, and positional encoding, which provided sequence order information. This architecture led to significant advancements in Natural Language Processing (NLP), enabling models like BERT (*19*) and GPT (*20*), which leveraged bidirectional and autoregressive pretraining respectively to further enhance sequence modeling capabilities.

Recent developments have extended Transformers with more efficient and flexible designs, particularly for handling sequence generation and long-range dependencies. One notable advancement is the Non-Autoregressive Transformer (NAT) (*21*), which eliminates the sequential dependency in decoding by generating all tokens simultaneously. This drastically improves efficiency compared to traditional autoregressive models but requires specialized methods such as iterative refinement or latent variable modeling to maintain generation quality. Stern et al. (2019) introduced Insertion Transformer (*22*), a non-autoregressive insertion-based process, allowing sequences to be generated in parallel by inserting tokens dynamically at appropriate positions rather than predicting them sequentially. Similarly, Gu et al. (2019) proposed Levenshtein Transformer (*23*) which leverages edit-based operations—insertions, deletions, and replacements—enabling more flexible and adaptive sequence generation, particularly useful for tasks requiring iterative refinement, such as text correction and sequence imputation.

**Location-Based Services (LBS) data**
Location-Based Service data (LBS) has emerged as a transformative resource for analyzing human mobility and travel behavior (*24*, *25*). The widespread adoption of smartphones, wearables, and positioning technologies enables the passive collection of vast amounts of high-frequency, granular geolocation information. This data typically includes timestamps and geographic coordinates, capturing individuals' movements and offering rich, dynamic insights into activity and travel patterns. Today, LBS plays a pivotal role in various applications, including transportation planning (*25*, *26*), crime prediction (*27*), public health (*28*, *29*), climate change analysis (*30*), and disaster management (*31*).

A key advantage of LBS data is its relatively low cost compared to traditional methods of collecting massive amount of human mobility data (*32*). Conventional approaches, such as household travel surveys, manual traffic counts, and GPS-based travel diaries, are often small-scale, labor-intensive, expensive and time-consuming. These methods typically require extensive logistical planning, participant recruitment, and rigorous data validation, all of which contribute to higher costs. In contrast, LBS data utilizes passively collected data from existing infrastructures like mobile apps, GPS systems, cellular networks, and Wi-Fi connections. This passive collection process eliminates the need for direct participant engagement, significantly reducing operational expenses.

However, the quality of LBS data, particularly its spatio-temporal density, critically impacts the accuracy of mobility inferences, such as identifying trips, stay points, and activity durations. As shown in **Figure 1**, a complete sighting dataset provides a wealth of spatial-temporal information, enabling more precise tracking of an individual's movements. Due to the high data density, as shown in the left subfigure, frequent sightings increase the reliability of inferring trips and activities from the sighting data. However, as shown in the right subfigure, the absence of some sightings can lead to missing certain trips or activities because it encompasses a wider range of potential travel paths and stopping places, making it difficult for algorithms to accurately infer movement patterns. Previous research has noted similar findings when inferring mobile metrics, such as stay points (*33*), trip rate, and dwell time (*34*), whereby high-quality sighting data, with more frequent and consistent records, enhances the reliability of inferring individual mobility metrics, while sparse or discontinuous data increases the bias of mobility metrics inference.

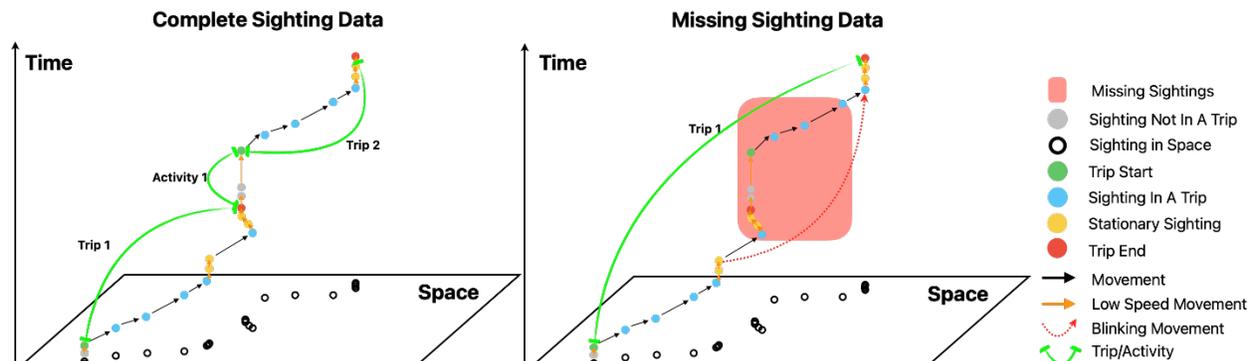

**Figure 1 Illustration of missing activities and trips due to missing sightings**

We employ a method to identify trips and tours from sightings in location-based service (LBS) data by using distance, speed, and time between sightings (*35*). In this study, the underlying assumption is that the LBS sighting data processed by this algorithm outputs meaningful trips, however, the inherent sparsity of LBS data often results in missing or incomplete sighting information, which omits some information from the identification trips. To overcome this, we leverage high-quality sighting data as a proxy for complete datasets, using it to train a model capable of restoring individual-level incomplete trip or activity sequences. In this way, we can recycle a large amount of valuable sparse data while compensating for the shortcomings of sparse data by recovering missing detailed activity patterns.

The central research problem we address is: *Can we use trip and activity sequence data derived from high-quality LBS data to recover incomplete trip or activity sequences?* This question is both novel and significant, as it marks the first time such an approach has been proposed in the field. One key challenge is simultaneously predicting activities and their sequence positions. The model must not only identify what activities are missing (e.g., work or go shopping) but also place them correctly within an existing sequence (e.g., insert work after home on weekdays, or insert recreation after home on weekends). This requires capturing temporal, spatial and even personal preference dependencies. Another challenge is the event-driven nature of activity sequences. Unlike traditional time series forecasting, where time intervals are fixed and uniform, human activities have variable durations and irregular timing. By integrating high-quality sighting data with algorithmic processing, we aim to pioneer a solution that improves the reliability and utility of LBS-based trip/activity insights, opening up new avenues for research and applications of location-based analytics.

**DATASETS**
This study utilizes a trip identification algorithm (*35*), thereby generating trip/activity roster data from raw sighting data, which is processed as the primary dataset for modeling. A brief description of the datasets is given below:
- **LBS Raw Sighting Data**: Comprises all sightings of individuals observed in Philadelphia County during January 2020, including 1,733,424 unique device IDs and approximately 1.8 billion sightings, serving as a robust foundation for tracking movement.
- **Trip/Activity Roster Data**: This dataset is derived from raw sighting data. Data for a device on a given date will be retained if the device is observed at least 2 days per month and is observed for at least 16 unique hours during that date. Combined with inferred home and work locations, this dataset contains trips and activities, focusing on individuals with home addresses in Philadelphia County during January 2020, ultimately identifying 1,501,923 identified trips from 33,969 device IDs.
- **External Data:**
  - **Point of Interest (POI) information** (*36*) are attached to first daily trip and trip destinations to infer activities. To maximize the completeness of POI information attachment, the dataset includes POIs from the states of Pennsylvania, Maryland, New York, New Jersey, Delaware, Washington D.C., and Virginia. North American Industry Classification System (NAICS) codes, street addresses, and operating hours, etc. These NAICS codes are then used to classify inferred activities into different categories, facilitating a structured analysis of activity-travel patterns. Each NAICS code is mapped to one of nine predefined activity categories (**Table 1**).
  - **American Community Survey (ACS)** (*36*) data is processed to get census-tract level information about income, age, race and education.
- **Activity Sequence Data for Modeling**: The trip/activity roster data was further processed to form a dataset to be placed into the proposed model. First, the single-day activities of a device were converted into a sequence that serve as a sample in the dataset. In order to simulate the real-world situation where activities are missing, we strategically remove activities from the samples as model inputs and use the corresponding samples without removal as outputs to form a training dataset, with the aim of having the model learn as many activity-missing scenarios as possible. This process creates a training dataset with 1,145,840 samples. Second, a subset (0.3%) of the training dataset (3,437 samples) with is randomly selected to train the model.
- **Covariates for Modeling**: Additional information of the activity sequences was also processed as various sequences as covariates for each sample in the dataset. The additional information is extracted from trip/activity roster data and external data. See details in **Table 2**.

**Table 1 Relations between activity category and NAICS code**

| NAICS Codes | Activity Category | Count |
|---|---|---|
| **42, 44, 45** | GoShopping | 234358 |
| **51, 53** | Other | 133036 |
| **52, 54, 56, 81, 92** | PersonalBusiness | 312299 |
| **61** | GoToSchool | 36198 |
| **62** | Healthcare | 181699 |

| 71 | Recreation | 52143 |
|---|---|---|
| 72 | EatOut | 150089 |
| - | HomeActivity | - |
| - | WorkForPay | - |
| **In Total:** | | 1099822 |

**Table 2 Covariates used in VSNIT model**

| Time-dependent Covariates | Description | Type | Values/Range |
|---|---|---|---|
| Time of Day (Arrival) | 15-minute intervals | Categorical | 1-96 (represents 15-minute interval) |
| Time of Day (Departure) | 15-minute intervals | Categorical | 1-96 (represents 15-minute interval) |
| Travel Mode | Mode of transportation | Categorical | 5 modes + Unknown |
| Trip Distance | Distance of trip | Continuous | Miles |
| Weekday | Day of the week | Categorical | 1-7 (e.g., 1=Monday, 7=Sunday) |
| Is Holiday | Whether the day is a holiday | Categorical | Binary (0=Non-holiday, 1=Holiday) |
| **Static Covariates** | | | |
| Tract Income Category | Income level of census tract | Categorical | 1-5 (e.g., 1=Lowest, 5=Highest) |
| Tract Age Category | Age demographic of census tract | Categorical | 1-5 |
| Tract Race Category | Racial demographic of census tract | Categorical | 1-5 |
| Tract Education Category | Education level of census tract | Categorical | 1-5 |

**METHODS**

A new model structure: Variable Selection Network-fused Insertion Transformer (VSNIT) is introduced as an advanced computational framework designed to address the dual challenges of activity sequence recovery and generation. This architecture synergistically integrates the Insertion Transformer (*22*) and the Variable Selection Network (VSN), a component derived from the Temporal Fusion Transformer (TFT) (*37*), to provide a robust solution for processing complex, temporally structured activity sequence data. The performance of VSNIT is compared to a vanilla Insertion Transformer model (*22*) as baseline. Since vanilla Insertion Transformer was initially designed for machine translation task, the baseline model only takes activity sequence itself as input to recover the sequence.

The Insertion Transformer forms the backbone of VSNIT, offering unparalleled flexibility in sequence construction by permitting tokens—representing individual activities—to be inserted at any position within a sequence during the decoding phase. This capability is ideally suited to the activity sequence recovery task, where the *identified activities serve as the initial sequence and the model iteratively inserts arbitrary activity into any position in the sequence*. Unlike traditional autoregressive models that generate sequences strictly from left to right, the Insertion Transformer jointly models *what* to insert (i.e., the activity) and *where* to insert it (i.e., the position within the sequence). For recovery tasks,

VSNIT initializes with an incomplete sequence of inferred activities and iteratively performs insertion operations to reconstruct missing segments, ensuring that the integrity of the real data is preserved while completing the sequence in a contextually coherent manner.

The Variable Selection Network (VSN), adapted from the Temporal Fusion Transformer (TFT) (*37*), augments VSNIT by providing instance-wise selection of relevant input variables at each decoding step. Real-world activity datasets typically encompass a variety of features, including static covariates (e.g., sociodemographic attributes) and time-dependent covariates (e.g., time of day or travel mode). However, the relevance and predictive contribution of these variables are often unknown, and many may introduce noise that degrades model performance. VSN addresses this challenge by dynamically identifying and prioritizing the most salient variables—both static and time-dependent—for each activity or sequence segment.

**Problem statement**
This study tries to tackle the challenge of using a predictive model to recover incomplete daily activity sequences or generate complete sequences from scratch. Assuming a complete daily sequence is represented as:

$$Y_{i,d} = [y_{i,1}, y_{i,2}, \dots y_{i,n_{i,d}}]^T \in \mathbb{R}^{n_{i,d}} \quad (1)$$

In **Eq. (1)**, $y_{i,n_{i,d}}$ represents $n_{i,d}$-th activity of individual $i$ on day $d$, with the total number of activities $n_{i,d}$ varying across individuals and dates.

The model aims to both identify the missing activities and determine their appropriate locations within an incomplete or blank sequence for a given individual on a specific day. To achieve this, the study utilizes covariates:

$$X_{i,d} = [x_{i,1}, x_{i,2}, \dots x_{i,n_{i,d}}]^T \in \mathbb{R}^{n_{i,d} \times N_C} \quad (2)$$

**Eq. (2)** provides critical contextual input, categorized into three types:
1) static covariates $s_i \in \mathbb{R}^{d_s}$: demographics, socioeconomics, etc.
2) time-dependent known covariates $u_{i,t} \in \mathbb{R}^{d_u}$: weekday, holiday, etc.
3) time-dependent unknown covariates $v_{i,t} \in \mathbb{R}^{d_v}$: travel mode, trip distance, time of day, etc.

Since all covariates belong to the three types above, $d_s + d_u + d_v = N_C$.

An incomplete daily sequence refers to any subset of a complete sequence that preserves the order. The goal is to learn a non-linear mapping $F(.)$:

$$\widehat{Y}_{i,d} = F\left(Y_{i,d}^{\text{incomplete}}, X_{i,d}^{\text{incomplete}}; \theta\right) \quad (3)$$

for each individual $i$ on day $d$.

In real-world scenarios, if the activity sequence is incomplete, then the corresponding covariates are also incomplete at the same time, especially for the time-dependent unknown covariates. Therefore, to avoid information leakage during training, **Eq. (3)** can be rewritten as below:

$$\widehat{Y}_{i,d} = F\left(Y_{i,d}^{\text{incomplete}}, v_{i,d}^{\text{incomplete}}, u_{i,d}, s_{i,d}; \theta\right) \quad (4)$$

The primary goal is to leverage these diverse covariates to accurately recover missing parts of individuals' incomplete activity sequences or construct entirely new sequences.

Weiyu Luo, Chenfeng Xiong

**Figure 2** illustrates the problem across multiple scenarios, employing a visual representation to clarify the problem statement. Pink boxes denote static covariates, yellow boxes indicate time-dependent covariates and green boxes represent the activity sequences, with blue icons signifying identified activities within the input data and orange icons denoting the recovered or generated activities resulting from a model. It is noteworthy that the length of both the input and output sequences can be varied, and the recovered activities can be strategically inserted at any position within the input sequence to reconstruct or generate a complete daily activity profile.

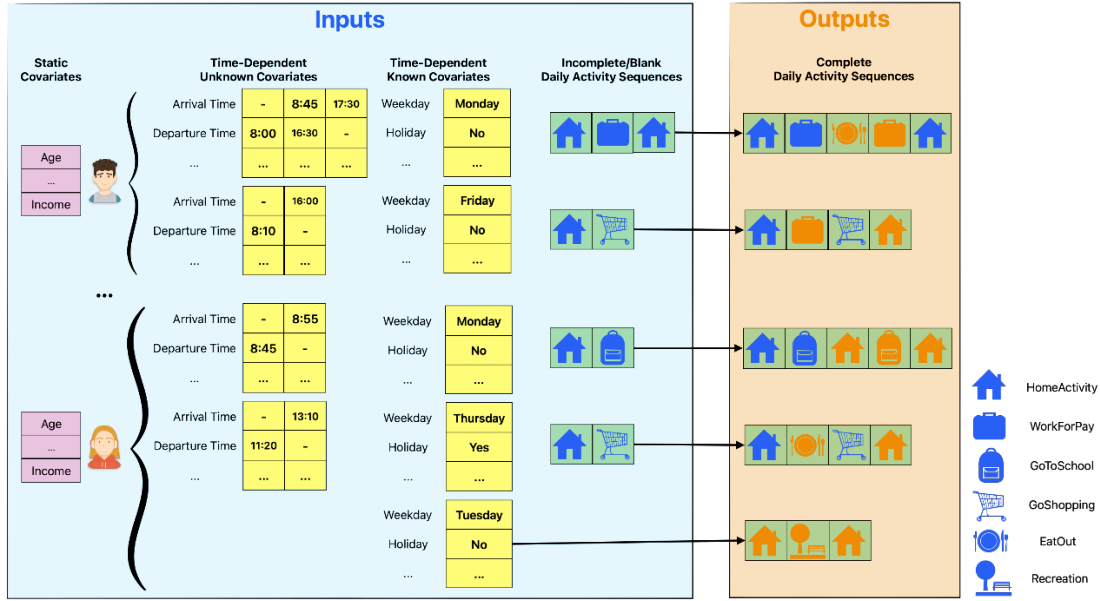

**Figure 2 Illustration of recovering incomplete daily activity sequences or generating new sequences from scratch**

**Model Structure and Components**
The VSNIT primarily takes inspiration from Insertion Transformer (*22*) and temporal fusion transformer (*37*). **Figure 3** demonstrates the architecture of VSNIT. Specifically, the VSNIT consists of these modules and sub-modules as follows:
1) Gated Residual Network (GRN). GRN serves as a building block of temporal fusion transformer (*37*). The GRN takes in a primary input $a$ and an optional context input vector $c$ and yields:

$$\mathrm{GRN}(a, c) = \mathrm{LayerNorm}\big(a + \mathrm{GLU}(\eta_1)\big) \tag{5}$$

$$\eta_1 = W_1 \eta_2 + b_1 \tag{6}$$

$$\eta_2 = \mathrm{ELU}(W_2 a + W_3 c + b_2) \tag{7}$$

where Exponential Linear Unit (ELU) is the Exponential Linear Unit activation function, $\eta_1, \eta_2 \in \mathbb{R}^{d_m}$ are intermediate layers. LayerNorm is standard layer normalization (*38*). ELU is formed as:

$$ELU(x) = \begin{cases} x, if\ x > 0 \\ \alpha(e^x - 1), if\ x \le 0 \end{cases} \tag{8}$$

where $\alpha = 1$. The Gated Linear Units (GLU) takes the form:

$$\text{GLU}(\gamma) = \sigma(W_4\gamma + b_4) \odot (W_5\gamma + b_5) \tag{9}$$

where $\sigma(.)$ is the sigmoid activation function, $W_{(.)} \in \mathbb{R}^{d_m \times d_m}, b_{(.)} \in \mathbb{R}^{d_m}$ are the weights and biases, $\odot$ is the element-wise Hadamard product. $d_m$ is the hidden state size common in VSN.

2) Variable Selection Network (VSN) (37). VSN takes time-dependent covariates as primary inputs and static covariates as context inputs. These covariates used in the model are shown in **Table 2**. Assuming $\xi_t^{(j)} \in \mathbb{R}^{d_m}$ denote the transformed input of the $j$-th time-dependent covariate at location $t$, with $\Xi_t = \left[\xi_t^{(1)T}, \ldots, \xi_t^{(m_\chi)T}\right]^T$ being the flatten vector of all inputs at location $t$.

$$v_{\chi t} = \text{Softmax}\left(\text{GRN}_{v_\chi}(\Xi_t, c_s)\right) \tag{10}$$

where $v_{\chi t} \in \mathbb{R}^{m_\chi}$ is a vector of variable selection weights, and $c_s$ are transformed inputs from static covariates. At each location, an additional layer of non-linear processing is employed by feeding each $\xi_t^{(j)}$ through its own GRN:

$$\tilde{\xi}_t^{(j)} = GRN_{\tilde{\xi}^{(j)}}\left(\xi_t^{(j)}\right) \tag{11}$$

$$\tilde{\xi}_t = \sum_{j=1}^{m_\chi} v_{\chi t}^{(j)} \tilde{\xi}_t^{(j)} \tag{12}$$

where $\tilde{\xi}_t^{(j)}$ is the processed feature vector for variable j. Note that each variable has its own $GRN_{\tilde{\xi}^{(j)}}$, with weights shared across all locations $t$.

3) Multi-head Attention: a self-attention mechanism is employed to learn long-term relationships across different locations in a sequence. Attention mechanisms scale values $V \in \mathbb{R}^{N \times d_V}$ based on relationships between keys $K \in \mathbb{R}^{N \times d_{attn}}$ and queries $Q \in \mathbb{R}^{N \times d_{attn}}$ as below:

$$\text{Attention}(Q, K, V) = \text{Softmax}\left(QK^T/\sqrt{d_{attn}}\right)V, \tag{13}$$

$$H_h = \text{Attention}\left(QW_Q^{(h)}, KW_K^{(h)}, VW_V^{(h)}\right), \tag{14}$$

$$\text{MultiHead}(Q, K, V) = [H_1, \ldots, H_{m_h}]W_H, \tag{15}$$

where $W_Q^{(h)} \in \mathbb{R}^{d_m \times d_{attn}}, W_K^{(h)} \in \mathbb{R}^{d_m \times d_{attn}}, W_V^{(h)} \in \mathbb{R}^{d_m \times d_V}$ are weights for each attention head for keys, queries and values. $W_H \in \mathbb{R}^{(m_h \cdot d_V) \times d_m}$ combines concatenated outputs from all heads $H_h$ linearly.

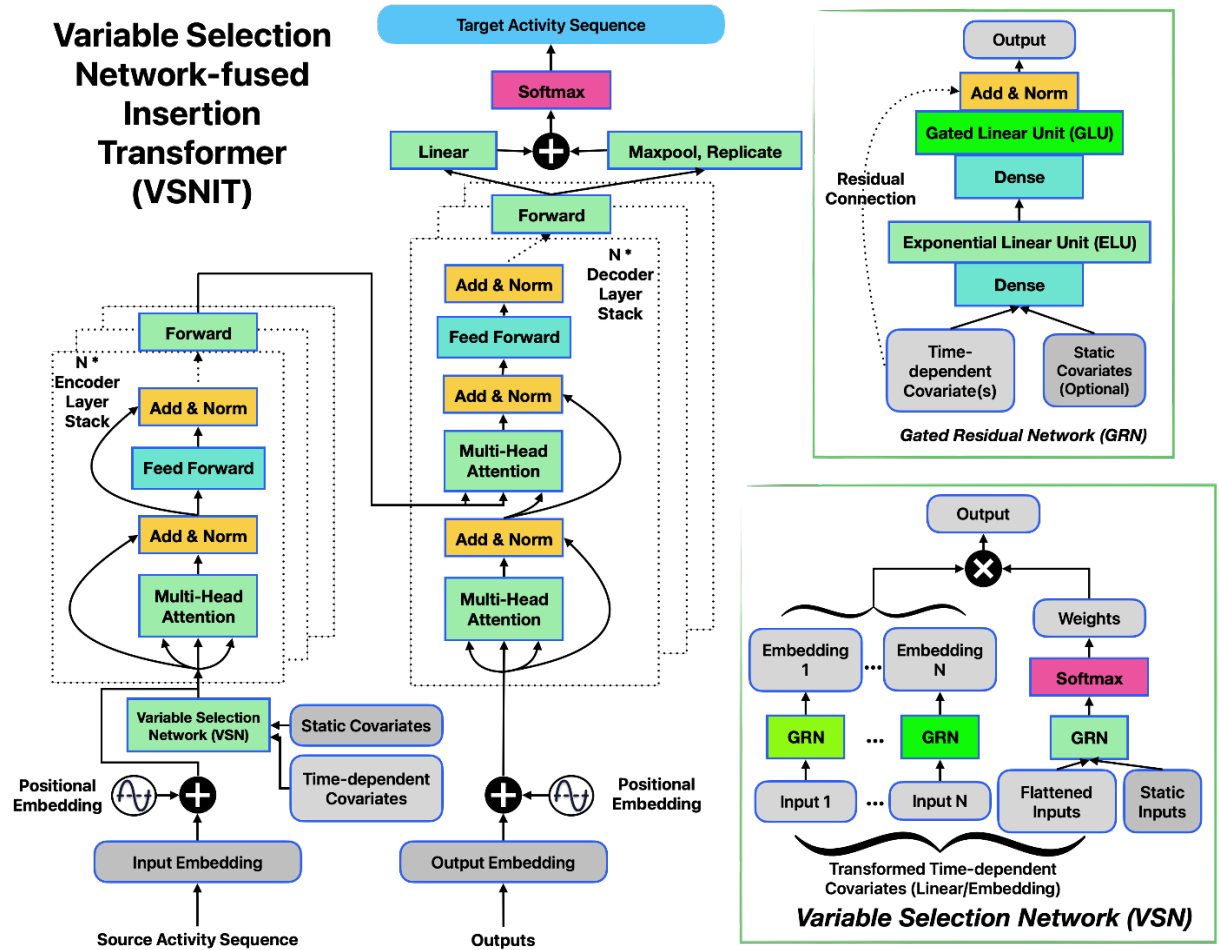

**Figure 3** Architecture of Variable-Selection-Network-fused Insertion Transformer (VSNIT)

### RESULTS
**Metrics**
**Table 3** shows the preliminary results comparing the proposed model to the baseline model. In the training dataset, a total of 5,757 activities is removed. In terms of the activity number in the recovered sequence, the proposed model significantly outperforms the baseline, achieving an average of 4.24 daily activities compared to the baseline's 3.43, with the target average being 4.49. This indicates a substantial improvement in the ability of the proposed model to recover a more accurate proportion of the original number of activities. On top of the improved accuracy of recovered activities, the proposed model also shows superior performance in terms of percentage metrics: it achieves an average percentage of correct insertion positions of 0.398 (0.396 for sequences with missing activities), compared to the baseline's 0.253 and 0.185, respectively, where "correct insertion position" means that the the inserted activity is placed in the correct position based on the anchoring of the source sequence, regardless of whether the activity itself is correct or not.

In addition, the proposed model correctly inserted 807 activities compared to only 459 in the baseline model, a notable achievement given that "correctly inserted activities" is the most stringent metric requiring both the prediction of the correct activity and the precise location of the activity in the sequence. Further analysis of the performance metrics shows that the proposed model has a precision of 0.165 (0.218 for the baseline model), a recall of 0.140 (0.080 for the baseline model), and an F1 score of 0.152 (0.117 for the baseline model), where precision is the ratio of correctly inserted activities to the total

Weiyu Luo, Chenfeng Xiong

number of inserted activities, recall is the ratio of correctly inserted activities to the total number of removed activities ratio, and the F1 score is the reconciled mean of precision and recall. Under "order-independent metrics," which assesses the model's ability to predict correct activities (regardless of their exact position in the sequence), the proposed model recovered 1,809 correct activities, with a precision of 0.370, a recall of 0.314, and an F1 score of 0.340, which is significantly more than the baseline's 925 correct activities, a precision of 0.439, a recall of 0.161, and an F1 score of 0.235, significantly exceeding the benchmark's 925 correct activities, 0.439 precision, 0.161 recall, and 0.235 F1 score. These results highlight the higher effectiveness of the proposed model in terms of overall activity recovery.

**Table 3 Precision, recall and F1-score of models**

| Metrics\Model | Proposed | Baseline |
|---|---|---|
| Total inserted activities (Total removed activities) | 4883(5757) | 2106(5757) |
| Average daily activities of hypothesis (target) | 4.24(4.49) | 3.43(4.49) |
| Average Correct inserted location percentage | 0.398 | 0.253 |
| Average Correct inserted location % (missing-token samples only) | 0.396 | 0.185 |
| Correct inserted activities | 807 | 459 |
| Precision (Correct inserted activities / Total inserted activities) | 0.165 | 0.218 |
| Recall (Correct inserted activities / Total removed activities) | 0.140 | 0.080 |
| F1 Score | 0.152 | 0.117 |
| **Order-independent Metrics** | | |
| Correct activities recovered | 1809 | 925 |
| Precision | 0.370 | 0.439 |
| Recall | 0.314 | 0.161 |
| F1 Score | 0.340 | 0.235 |

**Insertion patterns**

**Figure 4** compares the counts of the nine activities in the recovered sequences generated by the proposed model (hypothesis (proposed)) and the baseline model (hypothesis (proposed)) against the target values. The results show that the proposed model significantly outperforms the baseline model in terms of recovering the overall distribution of all nine activities, demonstrating a consistent improvement in reconstructing the activity distribution.

**Figure 5** shows the distribution of the top 20 inserted activities for the proposed and baseline models. The proposed model and the baseline model exhibit very different patterns of activity sequence recovery, possibly stemming from differences in their underlying architectures. The proposed model tends to insert more diverse and realistic activity patterns, capturing the variability inherent in human behavior more effectively by inserting over a wider range of activities. This suggests a smoother, more nuanced reconstruction of sequences that closely matches the natural heterogeneity observed in real-world activity data. In contrast, the baseline model tends to add no activity, or focuses insertions on a limited number of patterns, suggesting a bias toward oversimplified or repetitive structures. The baseline model's focus on a few major patterns may diminish its ability to reflect the complexity of the actual activity distribution, whereas the proposed model's smoother and more evenly distributed insertion strategy more realistically reflects the dynamic sequence of human activities.

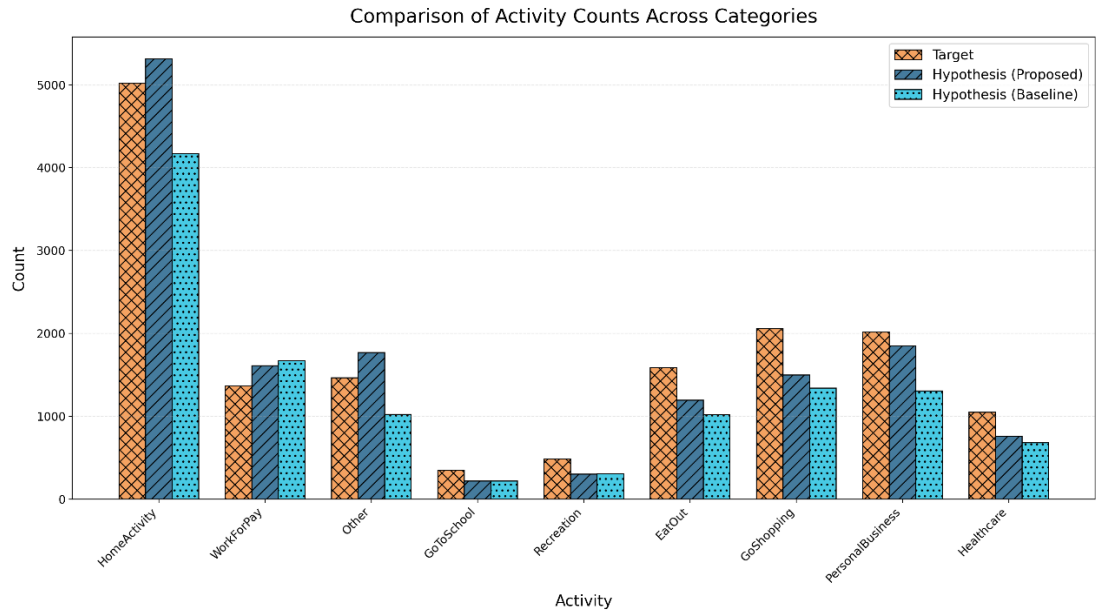

**Figure 4 Recovered activities distribution**

Weiyu Luo, Chenfeng Xiong

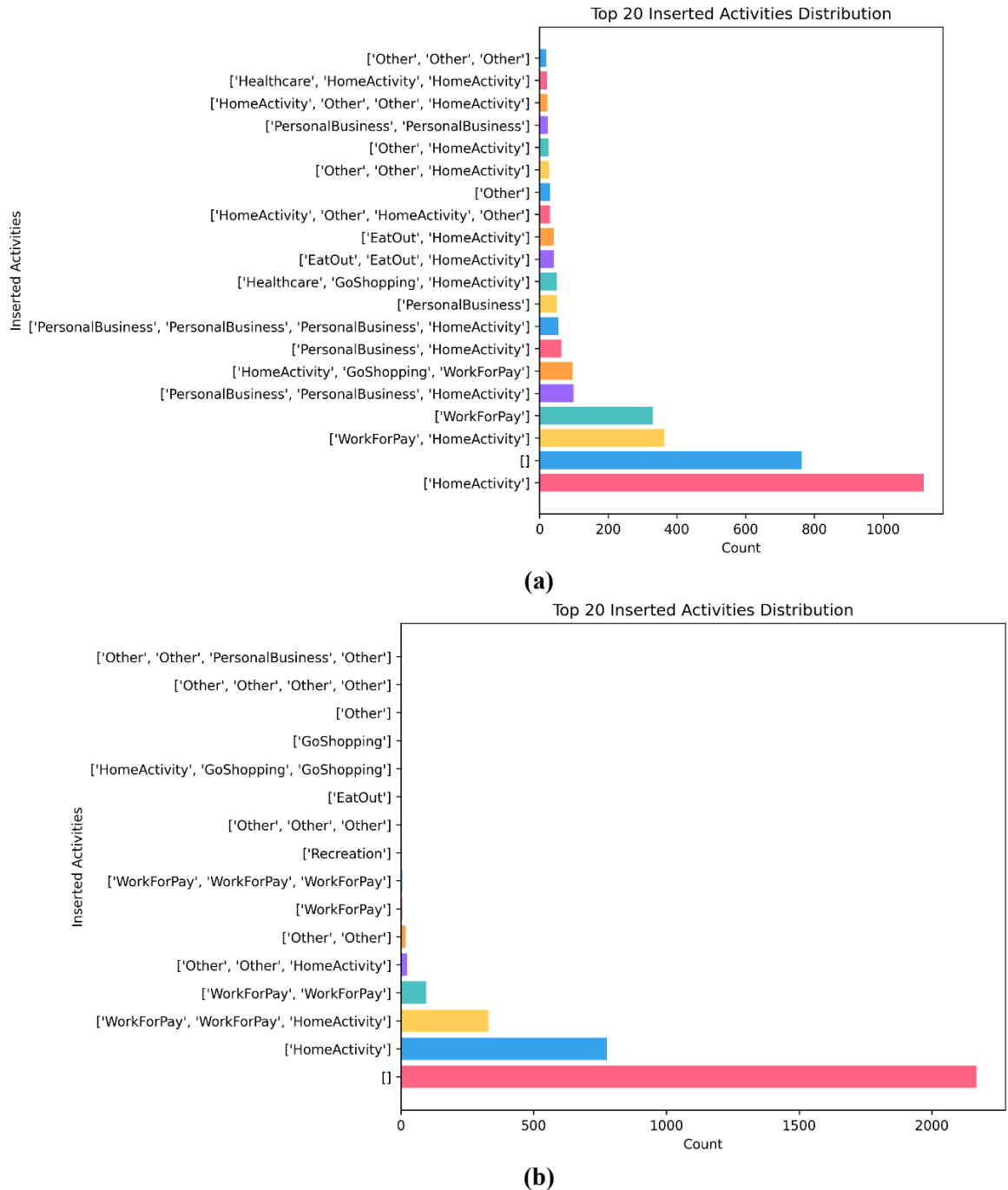

**Figure 5 Top 20 inserted activities distribution. (a) proposed model; (b) baseline model.**

In the LBS dataset, sighting data missing often lead to unrealistic activity transitions, as shown in **Figure 1**. When preprocessing the training data, this research strategically removes activities from the sequences to mimic the real-world situation. This can cause significant disruptions to the activity transitions of sequences, for example, deleting the activity "WorkForPay" in the sequence ["HomeActivity", "WorkForPay", "HomeActivity"] would result in a "HomeActivity" -> "WorkForPay" and a "WorkForPay" -> "HomeActivity" being corrupted. Conversely, inserting a "WorkForPay" activity into the sequence ["HomeActivity", "HomeActivity"] adds a "HomeActivity" -> "WorkForPay" and a "WorkForPay" -> "HomeActivity" transition.

**Figure 6** compares the counts of inserted transitions of the proposed model and of the baseline model against the counts of broken transitions in the preprocessing step. The number labeled "**" indicate closer to the broken transition count compared to the competing model. The left three figures represent counts where the inserted or removed activity serves as the second activity, and the right three figures where it acts as the first activity. As anticipated, the proposed model introduces more diverse modifications to activity transitions compared to the baseline. Among the 162 (81×2) transition types evaluated, the proposed model aligns more closely with the target in 48 cases, while the baseline outperforms in 26, with 88 resulting in ties. This suggests that in terms of the total number of transitions restored, across all broken transitions caused by activity removal, the proposed model more effectively recovers transitions through insertion. Furthermore, the results provide a more intuitive demonstration that the proposed model tends to insert a wider variety of transition patterns, enhancing the realism of the reconstructed sequences.

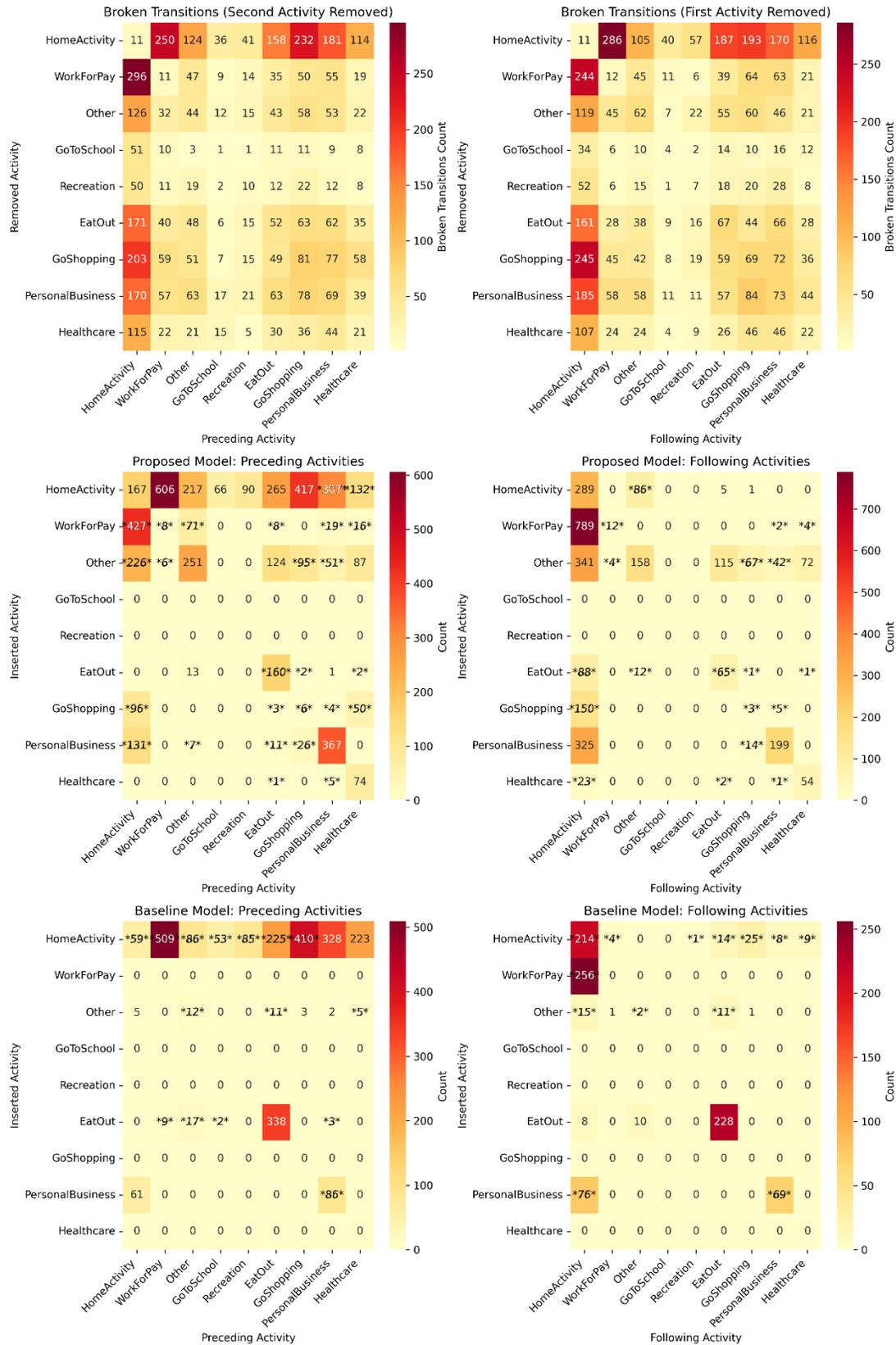

**Figure 6 Broken Transitions vs. Inserted Activity vs. Preceding/Following Activity.**

Weiyu Luo, Chenfeng Xiong

**DISCUSSION**

The results demonstrate VSNIT's superior performance over the baseline model. Firstly, in activity counts and their distribution, as shown in **Figure 5**, VSNIT outperforms the baseline by inserting activities more decisively—or "bravely"—to align with the target sequence, while also achieving a distribution that more closely mirrors the target dataset (**Figure 4**). This success can be attributed to two key factors: the integration of covariates and the VSN's advanced information extraction capabilities. By adaptively selecting the most relevant features (e.g., temporal or spatial context), VSNIT better captures the variability in human behavior, resulting in more accurate and representative activity insertions. In contrast, the baseline model, lacking such dynamic covariate handling, produces less diverse and less precise results. This enhancement underscores VSNIT's potential to maximize the value of sparse LBS data for mobility analysis.

In terms of activity transition distribution, as shown in **Figure 6**, both VSNIT and the baseline model capture some common patterns, such as returning home from other activities or transitioning to dining locations. However, VSNIT exhibits a broader and more nuanced learning capacity. It successfully learned additional patterns, including commuting to work from home or other locations and transitions between the three activities of "EatOut," "GoShopping," and "PersonalBusiness." This expanded pattern recognition reflects VSNIT's ability to model the complexity of real-world mobility more comprehensively. The model's covariate-driven approach likely enables it to discern subtle contextual cues—such as time of day or user demographics—that influence these transitions, allowing it to restore sequences with greater diversity and fidelity. These results affirm VSNIT's edge over existing methods and its suitability for tackling the intricacies of individual-level sequence recovery.

Although VSNIT performs well, its accuracy could be improved. There are several limitations of the current approach: first, the level of individual-level heterogeneity is still insufficient, and this can be improved by further feature engineering to better capture the nuances of individual behavior. Second, incorporating a wider range of data sources or finer-grained data, such as socio-demographic details at the census block level, weather, etc., could further enhance the model's context-awareness. Moreover, scaling up training with larger datasets could boost generalizability, enabling the model to handle edge cases and rare activity patterns more effectively. These improvements will solidify the reliability of VSNIT and expand its applicability in areas such as travel behavior analysis and transportation planning.

**CONCLUSIONS**

This study introduces the Variable Selection Network-fused Insertion Transformer (VSNIT) as a pioneering solution to the novel challenge of recovering incomplete activity sequences from sparse Location-Based Service (LBS) data. The results demonstrate that VSNIT significantly outperforms the baseline model across multiple dimensions: it inserts more diverse and realistic activity patterns, achieves higher accuracy in activity counts and distributions, and effectively restores disrupted transitions, particularly those involving home, work, shopping, and dining. These findings underscore the model's ability to capture the complexity and variability of human mobility, driven by its innovative integration of the Insertion Transformer's flexible sequence construction and the Variable Selection Network's dynamic covariate handling. Notably, VSNIT's great performance on a small dataset highlights its practical utility in real-world scenarios where data sparsity is a common challenge. Future research can further enhance the capabilities of VSNIT by integrating more data sources (e.g., weather or social media) and increasing the training data to improve the accuracy and robustness of its context-awareness and sequence recovery, enabling the model to be widely applied in different regions to replicate near-real human activities.